# Semi-Automatic Data Annotation, POS Tagging and Mildly Context-Sensitive Disambiguation:
# the eXtended Revised AraMorph (XRAM)

**Giuliano Lancioni, Valeria Pettinari, Laura Garofalo**
Department of Foreign Languages, Culture and Civilizations
Roma Tre University
via Ostiense 236,
Rome (Italy)
`giuliano.lancioni@uniroma3.it`
`pettinari.valeria@libero.it`
`laura.garofalo5@gmail.com`

**Marta Campanelli, Ivana Pepe, Simona Olivieri**
Department Italian Institute of Oriental Studies
Sapienza University of Rome
via Principe Amedeo, 182b
Rome (Italy)
`martac184@gmail.com`
`ivanapepe27@gmail.com`
`simolivieri@gmail.com`

**Ilaria Cicola**
EPHE
4-14 rue Ferrus, 75014
Paris (France)
Department Italian Institute of Oriental Studies, Sapienza University of Rome
Rome,
via Principe Amedeo, 182b
Rome (Italy)
`ilaria.cicola@etu.ephe.fr`
`ilaria.cicola@gmail.com`

## Abstract

An extended, revised form of Tim Buckwalter's Arabic lexical and morphological resource AraMorph, eXtended Revised AraMorph (henceforth XRAM), is presented which addresses a number of weaknesses and inconsistencies of the original model by allowing a wider coverage of real-world Classical and contemporary (both formal and informal) Arabic texts. Building upon previous research, XRAM enhancements include (i) flag-selectable usage markers, (ii) probabilistic mildly context-sensitive POS tagging, filtering, disambiguation and ranking of alternative morphological analyses, (iii) semi-automatic increment of lexical coverage through extraction of lexical and morphological information from existing lexical resources. Testing of XRAM through a front-end Python module showed a remarkable success level.

## 1 Introduction

Tim Buckwalter's AraMorph (henceforth AM: see Buckwalter, 2002) is one of the most widespread electronic resources for the Arabic lexicon and morphology. Applications using it include text analyzers, ontologies (e.g., Arabic WordNet browser, see Fellbaum et al., 2006), data mining and content extraction (e.g., ArMExLeR, Lancioni et al., 2013).

However, the original version of AM shows a number of shortcomings which reduce the coverage of the morphological analyzer and hinder its applicability to a number of genres and text types. In particular, Buckwalter (2002) focused mainly on contemporary newspaper texts, which makes the analyzer both underrecognize —because of lack of lexical and morphological coverage— and overrecognize (by spuriously increasing the amount of ambiguity because of the inclusion of historically and linguistically implausible alternatives) texts from other genres.

Some of these inconsistencies were tackled by the Revised AM model (henceforth RAM) presented in Boella et al. (2011). However, the necessity of a structural, opposite to incremental, revision and expansion of AM appears clearly in the impossibility to let a merely increased version effectively go beyond a certain level of performance in analyzing, e.g. Classical and modern informal texts.

XRAM presents itself as a structurally revised AM, which alters the basic original structure by adding usage and genre markers and by accruing the original, rigidly context-free conception of the analyzer by limited statistically gathered contextual selection information. These enhancements allow for a sensibly higher level of performance (see Section 3).

## 2 Description of XRAM

XRAM, just like AM and RAM, has a purpose of analyzing texts, but in a much more defined and thorough way.

In order to enhance the accuracy of the analysis we implemented a flag-selectable usage markers tool through the addition of a supplementary field in the Buckwalter analyzer (cfr infra § 2.1).

After selecting a single flag or a set of flags, according to the text genre, the text is tokenized and all the punctuation and formatting structure is stripped and factored out. Hence, the program produces a list of tokens ready to be processed by the XRAM analyzer, which aims to create a list of possible analyses for each token represented in the original text.

Types (distinct tokens) are analyzed and a dictionary of analysis, that assigns to each type a POS and a lemma, is created in order to reduce computing time.

As mentioned above, ambiguity is a significant weakness in the original AM model, which definitely compromises the correct analysis of the text. The XRAM RE module intervenes to reduce this ambiguity by filtering candidate analyses through a limited set of regular expressions. This module introduces a limited amount of context-sensitiveness in the system.

Analyses that survive the RE module are then ranked through a simple Language Model (LM) module, based upon Buckwalter and Parkinson's (2011) frequency list. Ranking introduces an order dimension in ambiguous analyses by assigning decreasing levels of plausibility to POS-lemma tuples.

XRAM capitalizes on the LM module by producing a semi-automated XML tagging of the original text according to the TEI P5 standard: the analysis with the higher rank is proposed as the default analysis, while other ones, lower in rank, are written in the XML output as alternative analyses.

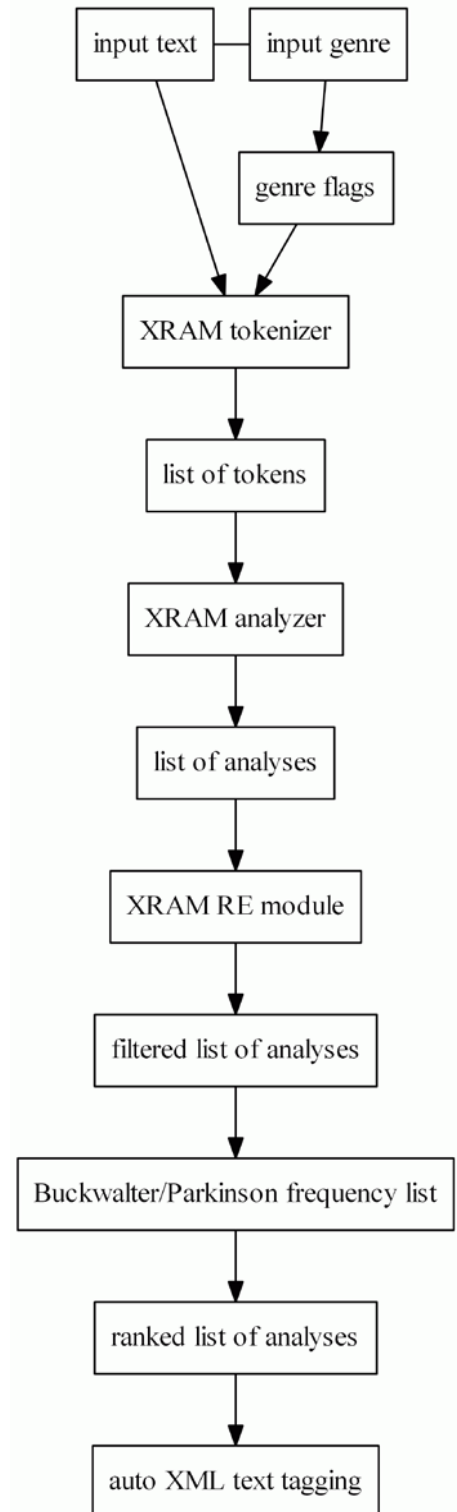

Figure 1: the XRAM pipeline

### 2.1 Flag-selectable usage markers

In order to make XRAM linguistic analysis even more reliable, markers are provided for graphemic, morphological and lexical features belonging to specific language varieties among Classical Arabic (CA), Modern Standard Arabic (MSA, formal), and Informal Colloquial Arabic

(ICA, informal) and technical and scientific sublanguage. Inspiration came from Buckwalter & Parkinson's (2011) Frequency Dictionary (Buckwalter and Parkinson, 2011): for each recorded lemma, the dictionary provides morphological, syntactic, orthographic and phonetic information as well as usage restrictions and register variations, according to the corpus where a lemma can be found, exclusively or most frequently.

Markers are encoded with flags which can be selected or unselected according to the language variety or genre the corpus to be processed is representative of. The genre selection step is user input by now, since it is outside the main task of the project, but several ways to detect it (semi-)automatically might be envisaged. This allows the analyzer reduce the amount of false positives by discarding of non-relevant gender- and variety-specific features. Flags were specified according to a number of diaphasic classification criteria, taking into account lexical expansion and morphological phenomena. Flags are labeled as follows:

| FLAG | FEATURE |
|---|---|
| XRAM_CA | Classical Arabic |
| XRAM_MSA | Modern Standard Arabic |
| XRAM_ICA | Informal Colloquial Arabic |
| XRAM_SPEC_MED | Medical Sublanguage |
| XRAM_SPEC_ALCH | Alchemic Sublanguage |
| XRAM_SPEC_GRAM | Grammatical Sublanguage |
| XRAM_NE | Name Entities |
| XRAM_FNE | Foreign Name Entities |
| XRAM_CAP | Colloquial Aspectual Preverbs |

Table 1: Genre Flags

Existing flags reflect the range of text genres included in the corpuses and subcorpuses available in our research. The system can be easily expanded by adding new flags.

Flags selection is usually compounded: for example, when processing a corpus of classical texts, XRAM_MSA, XRAM_ICA, XRAM_SPEC_MED, XRAM_FNE and XRAM_CAP flags will be deselected in order to optimize the output analysis.

Flags can be easily and efficiently implemented according to standard IT practices (as XORed bits), which makes genre and text type filtering quick and consistent.

## 2.2 Probabilistic mildly context-sensitive annotation

POS-tagging is one of the core tasks AM carries out since its inception, along with tokenization and word-segmentation. Since no syntactic information is provided to the program, AM shows a high degree of morphological and lexical ambiguity, particularly when processing unvocalized texts, due to the homography which characterizes written Arabic

```
e.g.
WORD:     الكتاب

Al+ktAb+
الكتاب
1    Al+kitAb+    ال*كتاب*
     kitAb_1      [كتب]
the+book+
Al/DET+Ndu+

2    Al+kut~Ab+   ال*كُتّاب*
     kut~Ab_1    [كتب]
the+kuttab (village
school);Quran school+
Al/DET+N+

3    Al+kut~Ab+   ال*كُتّاب*
     kAtib_1      [كتب]
the+authors;writers+
Al/DET+N+
```

To overcome this weakness, the Revised version of AraMorph, RAM (Boella et al., 2011) relayed on the vocalization of hadith texts. Notwithstanding, RAM produces good results only when processing a restricted range of text-genres, i.e. CA vocalized texts. This is why a further improvement of RAM is needed through the application of a mildly context-sensitive process of disambiguation. Specifically we adopted a streamline of two different but complementary approaches: (i) a filtering RE component (XRAM RE module) and (ii) a ranking LM module.

On the one hand, the filtering RE component reduces the amount of possible analyses by filtering out candidate sequences through regular expressions. E.g., the preposition مع *maʿa* 'with' unambiguously requires to be followed by a noun or (marginally) an adjective: the RE component includes a rule (symbolically represented as [* مع V]) to filter out candidate analyses of a word as a verb when preceded by مع.

On the other hand, the LM component ranks candidate analysis according to the probability of individual POS-lemma tuples. This is a local-sensitive disambiguation strategy which guides the ranking of alternative morphological analysis for each lemma identified by XRAM. The more these word combinations are likely to occur in the training and testing bases for this kind of strategy, the higher would be their ranking level, i.e. they will occupy top positions in the list of analyses provided by XRAM.

The LM component uses a hybrid approach: a order-3 language model drawn from a manually corrected sample is compounded with frequencies for individual POS-lemma tuples drawn from Buckwalter & Parkinson (2011).

This will drastically change the previous versions of AM, giving the research in the field of Arabic Corpus Linguistics and Arabic Computational Linguistics a whole new perspective and a even more functional degree of analysis, creating a morphologic-syntax interface.

### 2.3 Lexical and morphological XML tagging of texts

Aiming to analyze texts taken from Arabic corpuses, specific sections of the study have been conducted designing materials on the XML language, using the model available in TEI (Text Encoding Initiative) [1] P5 structure for textual annotation. Textual annotation is schemed adopting specific tags which help users identify precise information behind markers. The TEI standard, which has been chosen for its versatility and adaptability to various typologies of texts, fits well these specific purposes, even if adapted by validators from to time, depending on cases.

Morphological and lexical annotations are instead based on results given by RAM which provides a precise analysis of each occurring lemma, giving information in matter of stems, function of the word and a series of tags showing morphological features.

Combination of the two systems showed a remarkable success level, enabling readers to clearly identify every available information on given materials, including both textual and word-related (morphological and lexical) information. In fact, in addition to tags and basilar information, such a mixing provides general information which clearly identify the main features of the texts (such as the average length, frequency and occurrence of lemmas, identification of specific elements) just interpreting the combinations derived from the two overlapped patterns.

By way of partially re-writing and so extend RAM operating range, a further development will be then the semi-automatic annotation of XML texts modeled on TEI structure.

Thus, analyzing Arabic annotated texts employing RAM, the result will provide each word with all possible readings, giving specific information for every reading annotated.

Furthermore, splitting information deriving from RAM analysis, process of combination is refined by embedding data in the XML elements provided by standard TEI. In particular, the tag used to identify a word from the text is <w>, with an additional series of attributes as 'lemma' or 'type' to distinguish base-forms and specific functions.

The system automatically assigns the top ranked analysis selected by the LM component the <w> tag (which, as a container, cannot be embedded for one and the same input word), while marking less likely analyses with the annotation tag <note> with the analysis encoded in the 'ana' attribute in order to distinguish different readings of the same word.

While reviewing the XML output text, the annotator can reverse the default analysis by adding an attribute ed="correct" to one of the <note> elements. An XLST transformation takes care of promoting the marked analysis to <w> and to demote the corresponding <w> analysis to a <note> marker.

A sample derivation is shown for the preposition phrase مع كاتب 'with a writer'. The XRAM analyzer outputs one analysis for مع (the output by the XRAM system reformats a subsets of information in AM and is in the form vocalized_form/lemma/pos):

```
maEa/maEa_1/PREP
```

while three analyses are yielded for كاتب:

```
1) kAtab+a/kAtab_1/V
2) kAtib/kAtib_1/N
3) kAtib/kAtib_2/A
```

---

[1] http://www.tei-c.org/index.xml.

Analysis #1 is filtered out by RE rule (* مع V), while the LM component ranks #2 over #3. This is the result fragment in XML notation:

```
<w ana="maEa/maEa_1/PREP">
    مع
</w>
<w ana="kAtib/kAtib_1/N">
    <note ana="kAtib/kAtib_2/A"/>
    كاتب
</w>
```

The fragment shows the unique analysis for مع and the top ranked analysis for كاتب encoded in the 'ana' attribute of the <w> tag, while the alternative analysis for كاتب is encoded as a note. If the annotator does prefer one of the alternative analysis, (s)he adds the attribute 'ed="correct"' to it:

```
<w ana="kAtib/kAtib_1/N">
    <note ana="kAtib/kAtib_2/A"
ed="correct"/>
    كاتب
</w>
```

and launches the XLST transformation which reverses the selection:

```
<w ana="kAtib/kAtib_2/A">
    <note ana="kAtib/kAtib_1/N"/>
    كاتب
</w>
```

### 2.4 Semi-automatic increment of lexical coverage

One of the weak points of AraMorph is the limited range of text genres on which the resource was based: the lexicon files as well as the compatibility tables included in the program are mostly based on newspaper texts and other Modern Standard Arabic non-literary texts, which largely comprise the LDC Arabic corpus. The program is not only unbalanced and representative of a limited part of the Arabic vocabulary, but its look up lists lack of any stylistic and chronological information as well. Because of this, various problems can arise from the analysis of other textual genres, especially Classical and both contemporary formal and informal ones. Analyses conducted on Pre-islamic and Classical texts, such as Hadith texts (Boella et al., 2011) reveal that the main weak points of AM are:

(i) the rejection or the wrong analysis of words such as the 'ā- interrogative prefix, as well as imperative verbs that are not included in AM due to their rare occurrence in targeted AM texts. In addition, other errors that occur with classical Arabic corpora, especially pre-Islamic, concern broken plurals as well as certain verb stems (mainly of, maṣdars, participles, quadrilitteral verbs, iussive verbs, passives) which are either uncommon, as in the case of the quadrilitteral تخندذ (see Table 1), or are written in a nonstandard form not recognized by the analyzer, for example with the sukūn on the last letter. Note that when dealing with poetry there are other metrical phenomenon that are not recognized by the analyzer such as the 'alif or the yā' followed by the ha' at the end of the verse to create a rhyme (this was found when inserting the poetical works, or Dīwān, of the pre-Islamic poetess Al-Ḫansā' as a corpus in the analyzer);

(ii) the risk of false positives due to the presence of contemporary named entities inside the AM lexical lists, which are included in the search even when a classical text is analyzed (the same point has already been approached and partially overcome within in the above mentioned Boella et al., 2011).

On the other hand, for contemporary formal texts among newspapers and novels as well as contemporary informal texts such as blogs and social networks, one of the most important problems is the lack of a graphemic standardization of:

(iii) *transliterated foreign words* that nowadays Arabic borrows especially from English and arranges phonetically according to dialect and idiosyncratic varieties, which influence their transcription[2]. Among these types there are not only proper nouns of people and places but also commune nouns (for some examples see Table 1);

(iv) *dialect words* which are also exposed to a strong idiosyncratic variety when they are transcribed (for some examples see Table 1).

Thus, the XRAM project aims at enhancing the AM through the inclusion of additional lists of prefixes, stems and suffixes with the relative combination tables, in order to face points (i), (iii) and (iv). Several parts of the above mentioned lists will be automatically extracted from Arabic lexical resources currently available in XML format. For Classical texts, one of the most important resources is Salmoné's Arabic-English dictionary (1889), which is entirely encoded according to TEI standards and downloadable in a XML file. As for transliterated foreign words a solution is proposed by cross checking the concerned items with Arabic Wikipedia that is one of the largest online encyclopedia in existence and its large list of named entities has already inspired projects meant to potentiate and expand other Arabic lexical resources like Arabic WordNet (Alkhalifa and Rodriguez, 2010). Inside the XRAM project, the use of Arabic Wikipedia is finalized to align the transcription of foreign words and thus add them in the Buckwalter lists.

---
[2] As of matter of the Egyptian variety, Rosenbaum (2014) defines this linguistic phenomenon "Egyptianized English".

In regard of the most frequent unanalyzed dialect words, the solution is to manually set a list to include in AM since XML resources are not available at the moment.

| Classical Arabic | quadril "become evil" | تخنذذ |
| --- | --- | --- |
| | maṣdar III "thrust of the spear" | الطعان |
| | ašuǧāʿun/ā inter. + adj. "brave" | اَشُجَاعٌ |
| Transl. foreign named entities | Arizona | اريزونا |
| | Youtube | يوتوب |
| | Huffington | هفنجتون |
| Transl. foreign comm. nouns | aircraft | إيركرافت |
| | protocol | بروتكول |
| | the autobus | الأتوبيس |
| Dialect words | illī/ relative pron | اللي |
| | āntūn/ 2nd-people plural | آنتون |
| | dā/ m. s. dem. pro./ adj. | دا |

Table 2: Sample of unrecognized words in AM

## 3 Validation and research grounds

The evaluation of a tool such as XRAM involves some differences from standard evaluation methods in lemmatization and POS tagging tasks, first and foremost because the system outputs, on purpose, all available analyses and does not yield an analysis (e.g., thorough tentative reconstruction or error correction) where the analyzer has found none.

A first evaluation metrics is the rate of unrecognized words according to text genre (see Table 1 above):

| GENRE | % unknown XRAM | % unknown AM |
| --- | --- | --- |
| Classical Arabic | 3.4 | 12.4 |
| Modern Standard Arabic | 1.7 | 2.5 |
| Informal Colloquial Arabic | 7.6 | 18.5 |
| Medical Sublanguage | 1.3 | 7.5 |
| Alchemic Sublanguage | 3.5 | 14.2 |
| Grammatical Sublanguage | 2.7 | 8.6 |
| Named Entities | 6.5 | 7.6 |
| Foreing Named Entities | 14.3 | 15.6 |
| Colloquial Aspectual Preverbs | 6.7 | 23.4 |

Table 3: Comparison of recognition rates in XRAM and AM

While performance of XRAM is marginally better than AM in MSA texts, more specific genres show a remarkably higher performance, because of usage markers and increased coverage of the lexica.

Classical evaluation parameters change dramatically whether we take into consideration either the top-ranked analysis or any analysis output by the system:

| | Top-ranked Analysis | Any Analysis |
| --- | --- | --- |
| Error rate | 19.07 | 7.57 |
| Precision | 86.44 | 96.43 |
| Recall | 95.25 | 97.25 |
| F1 | 90.63 | 96.84 |

Table 4: Comparison of results

## 4 Conclusion

XRAM significantly enhances AM performances, especially for genre-specific texts. The model can be further enhanced by widening the filtering and ranking modules and by increasing the coverage of the lexicon, while keeping ambiguity low through a more and more refined assignment of usage markers.

A further development involves integrating current research on formal grammar (specifically, Combinatory Categorial Grammar, CCG: Steedman, 1996) within the ranking module.